\documentclass[runningheads]{llncs}
\usepackage{graphicx}
\usepackage{amsmath,amssymb} 
\usepackage{color}
\usepackage[width=122mm,left=12mm,paperwidth=146mm,height=193mm,top=12mm,paperheight=217mm]{geometry}
\usepackage{comment}
\usepackage{booktabs}
\usepackage{multirow}
\usepackage{float}
\usepackage{calc}
\usepackage{adjustbox}
\usepackage{pdfpages} 
\usepackage{tikz}
\usetikzlibrary{positioning}
\usepackage[subtle,bibbreaks=normal,mathspacing=normal,mathdisplays=normal]{savetrees}
\definecolor{newcolor}{rgb}{0,0.4,0}
\definecolor{oldcolor}{rgb}{0.7,0.7,0.7}
\definecolor{todocolor}{rgb}{1.0,0.5,0.5}

\newcommand{\fig}[1]{\mbox{Fig.\ \ref{#1}}}
\newcommand{\vect}[1]{\mathbf{#1}}

\newcommand{\tab}[1]{\mbox{Tab.~\ref{#1}}}
\newcommand{\etal}{\emph{et al.}}

\newcommand{\bb}[1]{\textbf{#1}}

\DeclareMathOperator*{\argmin}{arg\,min}
\DeclareMathOperator*{\softmax}{softmax}

\newcommand{\incgraphicsbrbox}[3]{%
  \begin{tikzpicture}%
    \node[inner sep=0pt] (mynode) {\adjustimage{#1}{#2}};%
    \node at (mynode.south east) [inner sep=0.25mm,scale=1,above left=0.15ex, fill=white, opacity=0.5, text opacity=1] {#3};%
  \end{tikzpicture}%
}

\graphicspath{{./images/}}

\begin{document}
\pagestyle{headings}
\mainmatter
\def\ECCV18SubNumber{0000}  

\title{DeepTAM: Deep Tracking and Mapping} 

\titlerunning{DeepTAM: Deep Tracking and Mapping}

\authorrunning{H. Zhou, B. Ummenhofer and T. Brox}

\author{Huizhong Zhou\textsuperscript{*}\quad Benjamin Ummenhofer\textsuperscript{*}\quad Thomas Brox}


\institute{University of Freiburg \\ 
    \{zhouh, ummenhof, brox\}@cs.uni-freiburg.de
}

\maketitle
\renewcommand*{\thefootnote}{\fnsymbol{footnote}}
\footnotetext[0]{\textsuperscript{*}Equal contribution}
\begin{abstract}
We present a system for keyframe-based dense camera tracking and depth map estimation that is entirely learned. 
For tracking, we estimate small pose increments between the current camera image and a synthetic viewpoint.
This significantly simplifies the learning problem and alleviates the dataset bias for camera motions.
Further, we show that generating a large number of pose hypotheses leads to more accurate predictions.	
For mapping, we accumulate information in a cost volume centered at the current depth estimate.
The mapping network then combines the cost volume and the keyframe image to update the depth prediction, thereby effectively making use of depth measurements and image-based priors.
Our approach yields state-of-the-art results with few images and is robust with respect to noisy camera poses.

We demonstrate that the performance of our 6~DOF tracking competes with RGB-D tracking algorithms. We compare favorably against strong classic and deep learning powered dense depth algorithms.

\end{abstract}

\section{Introduction}

In contrast to recognition, there is limited work on applying deep learning to camera tracking or 3D mapping tasks.
This is because, in contrast to recognition, the field of 3D mapping is already in possession of very good solutions.
Nonetheless, learning approaches have much to offer for camera tracking and 3D mapping.
On the limited number of subtasks, where deep learning has been applied, it has outperformed classical techniques:
on disparity estimation all leading approaches are based on deep networks, and the first work on dense motion stereo~\cite{ummenhofer_demon:_2017} immediately achieved state-of-the-art performance on this task.

In this work, we extend the domain of learning-based mapping approaches further towards full-scale SLAM systems.
We present a deep learning approach for the two most important components in visual SLAM: camera pose tracking, and dense mapping.

The main contribution of the paper is a learned tracking network and a mapping network, which generalize well to new datasets and outperform strong competing algorithms. This is achieved by the following key components:
\begin{itemize}
\item a tracking network architecture for incremental frame to keyframe tracking designed to reduce the dataset bias problem.
\item a multiple hypothesis approach for camera poses which leads to more accurate pose estimation.
\item a mapping network architecture that combines depth measurements with image-based priors, which is highly robust and yields accurate depth maps.
\item an efficient depth refinement strategy combining a network with the narrow band technique.
\end{itemize}

The most related classical approach is DTAM~\cite{newcombe_dtam_2011}, which stands for Dense Tracking And Mapping.
Conceptually we follow a very similar approach, except that we formulate it as a learning problem.
Consequently, we call our approach DeepTAM.

For tracking, DeepTAM uses a neural network for aligning the current camera image to a keyframe --color and depth image-- to infer the camera pose.
To this end, we use a small and fast stack of networks which implement a coarse-to-fine approach.
The network stack incrementally refines the estimated camera pose.
In each step we update a virtual keyframe, thereby improving convergence of the predicted camera pose.
This incremental formulation significantly simplifies the learning task and reduces the effects of dataset bias.
In addition, we show that generating a large number of hypotheses improves the pose accuracy.

Our mapping network is built upon the plane sweep stereo idea \cite{collins_space-sweep_1996}.
We first accumulate information from multiple images in a cost volume, then extract the depth map using a deep network by combining image-based priors with the accumulated depth measurements.
To further improve the depth prediction we append a network, which iteratively refines the prediction using a cost volume defined on a narrow band around the previous surface estimate. The obtained depth can be a valuable cue for many vision tasks, e.g. object localization~\cite{song_joint_2015,dhiman_continuous_2016}, scene understanding~\cite{gupta_2013,gupta_2015}, 
image dehazing~\cite{fattal_2008,derain_zhang_2018,dehaze_zhang_2018}.

As a learning approach, DeepTAM is very good at integrating various cues and learning implicit priors about the used camera. 
This is in contrast to classic approaches which fundamentally rely on handcrafted features like SIFT \cite{lowe_distinctive_2004} and photoconsistency maximization.
A well-known problem of learning-based approaches is overfitting, and we took special care in the design of the architecture and the definition of the learning problem so that the network cannot learn simple shortcuts that would not generalize.

As a consequence, DeepTAM generalizes well to new datasets and is the first learned approach with full 6 DOF keyframe pose tracking and dense mapping.
On standard benchmarks, it compares favorably to state-of-the-art RGB-D tracking, while using less data.
DeepTAM employs dense mapping that can process arbitrary many frames and runs at interactive frame rates.

\section{Related work}

The most related work is DTAM~\cite{newcombe_dtam_2011}.
We build on the same generic idea: drift-free camera pose tracking via a dense depth map towards a keyframe and aggregation of depth over time.
However, we use completely different technology to implement this concept.
In particular, both the tracking and the mapping are implemented by deep networks, which solely learn the task from data.

Most related with regard to the learning methodology is DeMoN~\cite{ummenhofer_demon:_2017}, which implements 6 DOF egomotion and depth estimation for two images as a learning problem.
In contrast to DeMoN, we process more than two images.
We avoid drift by the use of keyframes, and we can refine the depth map as more frames are coming in.

A few more works based on deep learning have appeared recently that have a weak connection to the present work.
Agrawal \etal~\cite{agrawal_learning_2015} trains a neural network to estimate the egomotion, which mainly serves as a supervision for feature learning. Kendall et al. \cite{kendall_geometric_2017} apply deep learning to the camera localization task and Valada \etal~\cite{valada_deep_2018} show that the visual localization and odometry can be solved jointly within one network.
DeepVO~\cite{wang_deepvo_2017} runs a deep network for visual odometry, i.e., regressing the egomotion between two frames.
There is no mapping part, and the egomotion estimation only works for environments seen during training.
Zhou~\etal~\cite{zhou_unsupervised_2017} presented a deep network for egomotion and depth estimation that can be trained with an unsupervised loss.
The approach uses two images for depth estimation during training.
However, it ignores the second image when estimating the depth at runtime, hence ignoring the motion parallax.
SfM-Net~\cite{vijayanarasimhan_sfm-net_2017}, too, uses unsupervised learning ideas, and (despite its title) does not use the motion parallax for depth estimation. UnDeepVO~\cite{li_undeepvo_2017} proposed egomotion estimation and depth estimation again based on an unsupervised loss.
All these works are like DeMoN limited to the joint processing of two frames and limited to the motions present in the datasets.

Training and experiments in most of these previous works~\cite{zhou_unsupervised_2017,wang_deepvo_2017,li_undeepvo_2017} focus on the KITTI dataset~\cite{geiger_are_2012}.
These driving scenarios mostly show 3 DOF motion in a plane, which is induced by a 2 DOF action space (accelerate/brake, steer left/steer right).
In particular the hard ambiguities between camera translation and rotation do not exist since the car cannot move sideward.
In contrast, the present work yields full 6 DOF pose tracking, can handle these ambiguities, and we evaluate on a 6 DOF benchmark.

We cannot cover the full literature on classical tracking and mapping techniques, but there are some related works besides DTAM~\cite{newcombe_dtam_2011}
that are worth mentioning.
LSD-SLAM~\cite{engel_lsd-slam_2014} is a state-of-the-art SLAM approach that uses direct measures for optimization.
It provides a full SLAM pipeline with loop closing.
In contrast to DTAM and our approach, LSD-SLAM only yields sparse depth estimates.
Engel~\etal~\cite{engel_direct_2018} propose a sparse direct approach.
They show that integrating a sophisticated model of the image formation process significantly improves the accuracy.
For our learning-based approach, accounting for the characteristics of the imaging process is covered by the training process.
Similarly, Kerl~\etal~\cite{kerl_robust_2013} carefully model the noise distribution to improve robustness.
Again, this comes for free in a learning-based approach.
CNN-SLAM~\cite{tateno_cnn-slam_2017} extends LSD-SLAM with single image depth maps.
In contrast to our approach, tracking and mapping are not coupled in a dense manner.
In particular, the tracking uses a semi-dense subset of the depth map.

\section{Tracking}
Given the current camera image $\vect I^C$ and a keyframe, which consists of an image $\vect I^K$ and an inverse depth map $\vect D^K$, we want to estimate the $4\times4$ transformation matrix $\vect T^{KC}$ that maps a point in the keyframe coordinate system to the coordinate system of the current camera frame.
The keyframe pose $\vect T^K$ and the current camera pose $\vect T^C$ are related by
\begin{equation}
	\vect T^C = \vect T^K \vect T^{KC}, \quad\text{with}\; \vect T^C, \vect T^K, \vect T^{KC} \in \vect{SE}(3).
\end{equation}

Learning to compute $\vect T^{KC}$ is related to finding 2D-3D correspondences between the current image $\vect I^C$ and the keyframe $(\vect I^K, \vect D^K)$.
It is well known that the correspondence problem can be solved more efficiently and reliably if pixel displacements between image pairs are small.
Since we want to track the current camera pose at interactive rates, we assume that a guess $\vect T^V$ close to $\vect T^C$ is available.
Similar to DTAM \cite{newcombe_dtam_2011}, we generate a virtual keyframe $(\vect I^V, \vect D^V)$ that shows the content of the keyframe $(\vect I^K, \vect D^K)$ from a viewpoint corresponding to $\vect T^V$. Instead of directly estimating $\vect T^{KC}$, we learn to predict the increment $\delta \vect T$, i.e., we write the current camera pose as
\begin{equation}
	\label{eq:inc_motion}
	\vect T^C = \vect T^V \delta \vect T.
\end{equation}
This effectively reduces the problem to learning the function $\delta \vect T = f(\vect I^C, \vect I^V, \vect D^V)$.
We use a deep network to learn $f$.

\subsection{Network Architecture}
\begin{figure}
\centering
\adjincludegraphics[width=0.95\textwidth]{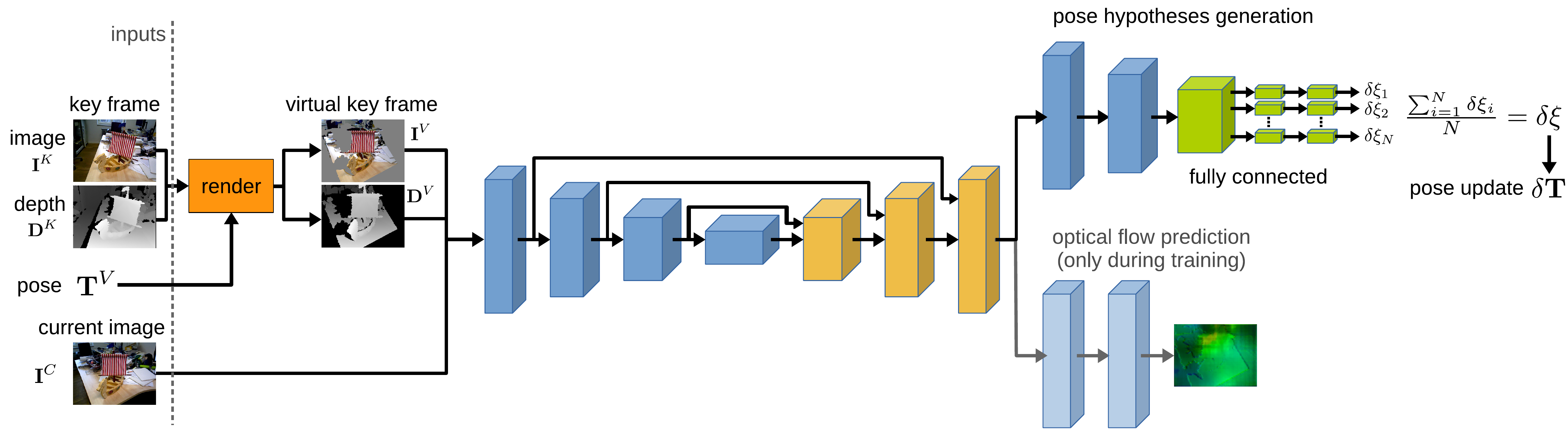}%
\caption{
The tracking network uses an encoder-decoder type architecture with direct connections between the encoding and decoding part.
The decoder is used by two tasks, which are optical flow prediction and the generation of pose hypotheses.
The optical flow prediction is a small stack of two convolution layers and is only active during training to stimulate the generation of motion features.
The pose hypotheses generation part is a stack of downsampling convolution layers followed by a fully connected layer, which then splits into $N=64$ fully connected branches sharing parameters to estimate the $\delta \xi_i$.
Along with the current camera image $\vect I^C$ we provide a virtual keyframe $(\vect I^V, \vect D^V)$ as input for the network, which is rendered using the active keyframe $(\vect I^K, \vect D^K)$ and the current pose estimate $\vect T^V$.
We stack the depicted network architecture three times with each instance operating at a different resolution as shown in \fig{fig:tracking_architecture}.
}
\label{fig:tracking_architecture_detail}
\end{figure}

We use the encoder-decoder-based architecture shown in \fig{fig:tracking_architecture_detail} for learning to estimate the 6 DOF pose between a keyframe $(\vect I^K, \vect D^K)$ and an image $\vect I^C$.
A detailed description of all network parameters can be found in the supplementary material.

Since camera motion can only be estimated by relating the keyframe to the current image, we make use of optical flow as an auxiliary task.
The predicted optical flow ensures that the network learns to exploit the relationship between both frames. We demonstrate the importance of the flow prediction in \tab{tab:tracking_error}.
We use the features shared with the optical flow prediction task in a second network branch for generating pose hypotheses.
As we show in the experiments (\tab{tab:tracking_error}), generating multiple hypotheses improves the accuracy of the predicted pose compared to the direct prediction of the pose.

The last part of the pose generation consists of $N=64$ branches of stacked, fully connected layers sharing their weights.
We found that this configuration is more stable and accurate than a single branch of fully connected layers computing $N$ poses.
Each generated pose hypothesis is a 6D pose vector $\delta \vect \xi_i = (\vect r_i, \vect t_i)^\top$.
The 3D rotation vector $\vect r_i$ is a minimal angle-axis representation with the angle encoded as the magnitude of the vector.
The translation $\vect t_i$ is encoded in 3D Cartesian coordinates.
For simplicity, and because $\delta \vect \xi_i$ are small rigid body motions, we compute the final pose estimate $\delta \vect \xi$ as the linear combination
\begin{equation}
\label{eq:mean_motion_prediction}
	\delta \vect \xi = \frac{1}{N} \sum_{i=1}^{N=64} \delta \vect \xi_i.
\end{equation}

\begin{figure}
\centering
\adjincludegraphics[width=0.95\textwidth]{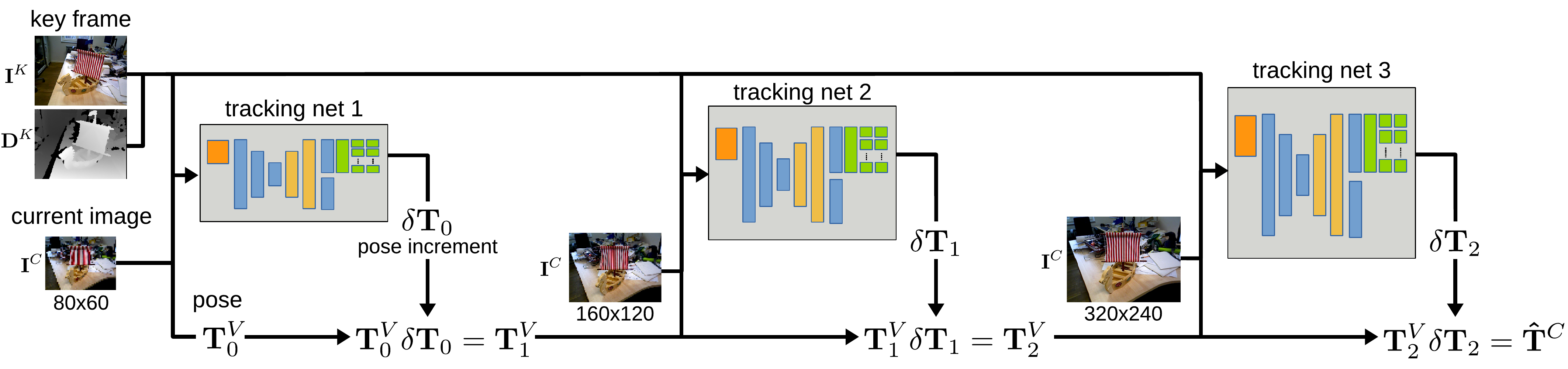}\\
\caption{
Overview of the tracking networks and the incremental pose estimation.
We apply a coarse-to-fine approach to efficiently estimate the current camera pose.
We train three tracking networks each specialized for a distinct resolution level corresponding to the input image dimensions $(80\times60)$, $(160\times120)$ and $(320\times240)$.
Each network computes a pose estimate $\delta \vect T_i$ with respect to a guess $\vect T^V_i$.
The guess $\vect T^V_0$ is the camera pose from the previously tracked frame.
Each of the tracking networks uses the latest pose guess to generate a virtual keyframe at the respective resolution level and thereby indirectly tracking the camera with respect to the original keyframe $(\vect I^K, \vect D^K)$.
The final pose estimate $\vect{\hat T}^C$ is computed as the product of all incremental pose updates $\delta \vect T_i$.
}
\label{fig:tracking_architecture}
\end{figure}

Coarse camera motions are already visible at small image resolutions, while small motions require higher image resolutions.
Thus, we use a coarse-to-fine strategy to efficiently track the camera in real time.
We train three distinct tracking networks as shown in \fig{fig:tracking_architecture}, which deal with the pose estimation problem at different resolutions and refine the prediction of the respective previous resolution level.

\subsection{Training}
A major problem of learning-based approaches is the strong dependency on suitable datasets.
Datasets often do not cover all important modes, which complicates generalization to new data.
An example is the KITTI dataset for autonomous driving \cite{geiger_are_2012}, which is limited to motion in a plane and does not cover full 6 DOF motion.
As a consequence, learning-based methods easily overfit to this type of motion and do not generalize.
Artificial data can be used to alleviate this problem, but it is not trivial to generate realistic imagery with ground truth depth. 

We tackle this problem in two ways. First by using the incremental formulation in \eqref{eq:inc_motion}, i.e., we estimate a small increment $\delta \vect T$ instead of the absolute motion between keyframe and current camera image.
This reduces the magnitude of motion and reduces the difficulty of the task.
Second, we use rendered images and depth maps as a proxy for real keyframes.
Given a keyframe $(\vect I^K, \vect D^K)$, we sample the initial pose guess $\vect T^V_0$ from a normal distribution centered at the ground truth pose $\vect T^C$ to generate the virtual frame $(\vect I^V, \vect D^V)$.
This simulates all possible 6 DOF motions and, thus, effectively augments the data to overcome the limited set of motions in the dataset.

\subsubsection{Datasets}
We train on image pairs from the SUN3D dataset \cite{xiao_sun3d_2013} and the SUNCG dataset \cite{song_semantic_2017}.
For SUN3D we sample image pairs with a baseline of up to 40cm.
For SUNCG we generate images with normally distributed baselines with standard deviation 15cm and rotation angles with standard deviation $0.15$ radians.
When sampling an image pair we reject samples with an image overlap of less than 50\%.
For keyframe depth maps $\vect D^K$, we use the ground truth depth from the datasets during training.

\subsubsection{Training Objective}
The objective function for the tracking network is
\begin{equation}
	\mathcal{L}_\text{tracking} = \mathcal{L}_\text{flow}(\vect w) + \mathcal{L}_\text{motion}\vect(\delta \vect \xi) + \mathcal{L}_\text{uncertainty}(\delta \vect \xi_i).
\end{equation}
The predicted optical flow $\vect w$ and the predicted poses $\delta \vect \xi_i$ are the network's outputs.

The loss $\mathcal{L}_\text{flow}$ defines the auxiliary optical flow task. We use the endpoint error
\begin{equation}
	\mathcal{L}_\text{flow} = \sum_{i,j} \left\Vert \vect w(i,j) - \vect w_\text{gt}(i,j) \right\Vert_2,
\end{equation}
which is a common error metric for optical flow.

The two losses $\mathcal{L}_\text{motion}$ and $\mathcal{L}_\text{uncertainty}$ for the generation of pose hypotheses are defined as:
\begin{align}
		\mathcal{L}_\text{motion} &= \alpha \left\Vert \vect r - \vect r_\text{gt} \right\Vert_2 + \left\Vert \vect t - \vect t_\text{gt} \right\Vert_2,\, \text{and} \\
\label{eq:loss_uncertainty}
		\mathcal{L}_\text{uncertainty} &= \frac{1}{2}\log\left(\vert\vect \Sigma\vert\right) - 2 \log\left( \frac{\vect x^\top \vect \Sigma^{-1} \vect x}{2} \right) - \log\left(K_v\left(\sqrt{2 \vect x^\top \vect \Sigma^{-1} \vect x }\right)\right).
\end{align}
The vectors $\vect r$ and $\vect t$ are the rotation and translation parts of the linear combination $\delta \vect \xi$ defined in \eqref{eq:mean_motion_prediction}.
We use the parameter $\alpha$ to balance the importance of both components.
We combine this loss, which directly acts on the predicted average motion, with $\mathcal{L}_\text{uncertainty}$, which is the negative log-likelihood of the multivariate Laplace distribution.
We compute $\vect \Sigma$ from the predicted pose samples as $\vect \Sigma = \frac{1}{N} \sum_i^N(\delta \vect \xi_i - \delta \vect \xi)(\delta \vect \xi_i - \delta \vect \xi)^\top$, and the vector $\vect x$ as $\vect x = \delta \vect \xi - \delta \vect \xi_\text{gt}$.
During optimization we treat $\vect x$ as a constant.
The function $K_v$ is the modified Bessel function of the second kind.
We empirically found that a loss based on the multivariate Laplace distribution yields better results than the multivariate Normal distribution.
The uncertainty loss pushes the network to predict distinct poses $\delta \vect \xi_i$.

We optimize using Adam \cite{kingma_adam_2014} with the learning rate schedule proposed in \cite{loshchilov_sgdr_2016}.
We implement and train the networks with Tensorflow~\cite{tensorflow2015-whitepaper}.
Training the tracking network takes less than a day on an NVIDIA GTX1080Ti.
We provide the detailed training parameters in the supplementary material.

\section{Mapping}
We describe the geometry of a scene as a set of depth maps, which we compute for every keyframe.
To achieve high-quality depth maps we accumulate information from multiple images in a cost volume.
The depth map is then extracted from the cost volume by means of a convolutional neural network.

Let $\vect C$ be the cost volume and $\vect C(\vect x, d)$ the photoconsistency cost for a pixel $\vect x$ at depth label $d \in B_\text{fb}$.
We define the set of $N$ depth labels for a fixed range $[d_\text{min}, d_\text{max}]$ as
\begin{equation}\label{eq:fb}
B_\text{fb} = \{ b_i | b_i = d_\text{min} + i \cdot \tfrac{d_\text{max}-d_\text{min}}{N-1}, i= 0, 1, ..., N-1\}.
\end{equation}

Given a sequence of $m$ images $\vect I_1,.., \vect I_{m}$ along with their camera poses $\vect T_1,.., \vect T_{m}$, we compute the photoconsistency costs as
\begin{equation}\label{eq:cv}
	\vect C(\vect x, d)= \sum_{i \in \{1,..,m\}}  \rho_i(\vect x, d) \cdot w_i(\vect x).
\end{equation}
The photoconsistency $\rho_i(\vect x, d)$ is the sum of absolute differences (SAD) of $3\times\!3$ patches between the keyframe image $\vect I^K$ and the warped image $\vect{\tilde I}_i$ at point $\vect x$ for depth $d$. We obtain $\vect{\tilde I}_i$ using a warping function $\mathcal W(\vect I_i, \vect T_i(\vect T^K)^{-1}, d)$, which warps the image $\vect I_i$ to the keyframe using the relative pose and the depth.

The weighting factor $w_i$ is then computed as
\begin{equation}
	\begin{aligned}
		w_i(\vect x) = 1- \frac{1}{N-1} \sum_{d \in B_\text{fb} \setminus \{d^*\}} {\rm exp} \left(-\alpha \cdot \left(\rho_i(\vect x, d) - \rho_i(\vect x, d^*)\right)^2 \right).
	\end{aligned}
\end{equation}
$w_i$ describes the matching confidence and is close to $1$ if there is a clear and unique minimum $\rho_i(\vect x, d^*)$ with $d^* = \argmin_d \rho_i(\vect x,d)$.

In classic methods the cost volume is taken as data term and a depth map can be obtained by searching for the minimum cost.
However, due to noise in the cost volume, various sophisticated regularization terms and optimization techniques have been introduced \cite{hirschmuller_accurate_2005,felzenszwalb_efficient_2006,hosni_fast_2013} to extract the depth in a robust manner.
Instead, we train a network to use the matching cost information in the cost volume and simultaneously combine it with the image-based scene priors to obtain more accurate and more robust depth estimates.

\begin{figure}
\centering
\adjincludegraphics[width=1.0\textwidth]{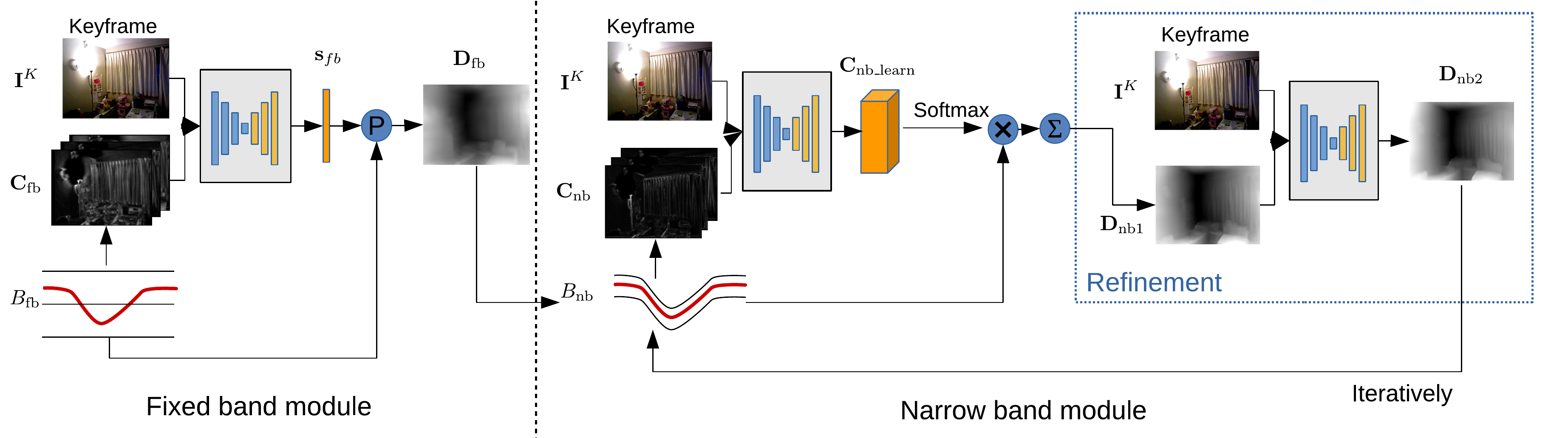}
\caption{Mapping networks overview. Mapping consists of a fixed band module and a narrow band module, which is based on an encoder-decoder architecture. \bb{Fixed band module}: This module takes the keyframe image $\vect I^K$ ($320 \times 240 \times 3$) and the cost volume $\vect C_{fb}$ ($320 \times 240 \times 32$)  generated with 32 depth labels equally spaced in the range [0.01, 2.5] as inputs and outputs an interpolation factor  $s_{fb}$ ($320 \times 240 \times 1$). The fixed band depth estimation is computed as $\vect D_{fb} = (1-\vect s_{fb}) \cdot d_{min} + \vect s_{fb} \cdot d_{max}$. \bb{Narrow band module}: The narrow band module is run iteratively; in each iteration we build a cost volume $\vect C_{nb}$ from a set of depth labels distributed around the current depth estimation with a band width $\sigma_{nb}$ of $0.0125$. It consists of two encoder-decoder pairs. The first pair gets the cost volume $\vect C_{nb}$ ($320 \times 240 \times 32$) and the keyframe image  $\vect I^K$ ($320 \times 240 \times 3$) as inputs and generates a learned cost volume $\vect C_{nb\_learn}$ ($320 \times 240 \times 32$). The depth map is then obtained using a differentiable soft argmin operation \cite{kendall_end--end_2017}: $\vect D_{nb1} = \sum_{d \in \vect B_{nb}} \vect B_{nb} \times \softmax(-\vect C_{nb\_learn})$. The second encoder-decoder pair gets the current depth estimation $\vect D_{nb1}$ and the keyframe image $\vect I^K$ and produces a refined depth $\vect D_{nb2}$.}
\label{fig:mapping_overview}
\end{figure}

For cost-volume-based methods, accuracy is limited by the number of depth labels $N$.
Hence, we use an adaptive narrow band strategy to increase the sampling density while keeping the number of labels constant.
We define the narrow band of depth labels centered at the previous depth estimate $d_\text{prev}$ as
\begin{equation}\label{eq:nb}
B_\text{nb} = \{ b_i | b_i = d_\text{prev} + i \cdot \sigma_\text{nb} \cdot d_\text{prev}, i= -\tfrac{N}{2}, ..., \tfrac{N-2}{2}\}.
\end{equation}
$\sigma_\text{nb}$ determines the narrow band width.
We recompute the cost volume for the narrow band for a small selection of frames and search again for a better depth estimate.
The narrow band allows us to recover more details in the depth map, but also requires a good initialization and regularization to keep the band in the right place. 
We address these tasks using multiple encoder-decoder type networks.
\fig{fig:mapping_overview} shows an overview of the mapping architecture with the fixed band and narrow band stage.


\subsection{Network Architecture}
The network is trained to predict the keyframe inverse depth $\vect D^K$ from the keyframe image $\vect I^K$ and the cost volume $\vect C$ computed from a set of images $\vect I_1, ..., \vect I_m$ and camera poses $\vect T_1, ..., \vect T_m$.
$\vect D^K$ is represented as inverse depth, which enables a more precise representation with closer distance.
We apply a coarse-to-fine strategy along the depth axis.
Thus, the mapping is divided into a fixed band module and a narrow band module.
The fixed band module builds the cost volume $\vect C_{fb}$ with depth labels evenly spaced in the whole depth range, while the narrow band cost volume $\vect C_{nb}$ centers at the current depth estimation and accumulates information in a small band close to the estimate.

\begin{figure}
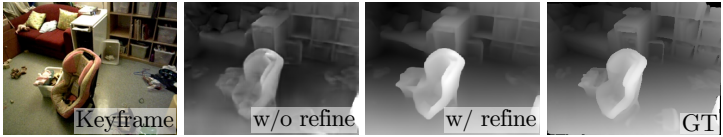

  \centering
    \incgraphicsbrbox{width=0.19\textwidth}{mapping/band/band_img.png}{Keyframe}
    \incgraphicsbrbox{width=0.19\textwidth}{mapping/band/band_only_prs_nb.png}{w\scalebox{0.9}{/}o refine}
    \incgraphicsbrbox{width=0.19\textwidth}{mapping/band/band_prs_nb.png}{w\scalebox{0.9}{/} refine}
    \incgraphicsbrbox{width=0.19\textwidth}{mapping/band/band_gt.png}{GT}
  \caption{Effects of the narrow band refinement. We apply the narrow band module for 15 iterations with and without refinement. 
Without the refinement, the module lacks the knowledge of the band shape and it can only make updates based on the measurements in the cost volume. 
This can help in capturing more details, but also causes strong artifacts.
Appending a refinement network with previous depth estimation as input allows for a better regularized and more stable depth estimation.
}
  \label{fig:mapping_refine}
\end{figure}

The fixed band module regresses an interpolation factor between the minimum and maximum depth label as output.
As a consequence, the network cannot reason about the absolute scale of the scenes, which helps to make the network more flexible and generalize better.
Unlike the fixed band, which contains a set of fronto-parallel planes as depth labels, the discrete labels of the narrow band  are individual for each pixel.
Predicting interpolation factors is not appropriate since the network in the narrow band module has no knowledge of the band's shape. 
We intentionally do not provide the narrow band network with the band shape (i.e. the depth value for which each depth label stands), because the network tends to overfit to this straight-forward cue and ignores the cost information in the cost volume.
However, the absence of the band shape makes the depth regularization difficult which can be observed in \fig{fig:mapping_refine}.
Therefore we append another refine network, which focuses on the problem of depth regularization.
Both networks together can be understood as solving alternatingly the data and smoothness terms of a variational approach.
The detailed architecture is shown in \fig{fig:mapping_overview}.

\subsection{Training}
We train our mapping networks from scratch using Adam \cite{kingma_adam_2014} based on the Tensorflow \cite{tensorflow2015-whitepaper} framework.
Our training procedure consists of multiple stages.
We first train the fixed band module with subsampled video sequences of length 8.
Then we fix the parameters and sequentially add the two narrow band encoder-decoder pairs to the training.
In the last stage we unroll the narrow band network to simulate 3 iterations and train all parts jointly.
Training the mapping networks takes about 8 days in total on an NVIDIA GTX 1080Ti.

\subsubsection{Datasets}
We train our mapping networks on various datasets to avoid overfitting. SUN3D \cite{xiao_sun3d_2013} has a large variety of indoor scenes.
For ground truth we take the improved Kinect depths with multi-frame TSDF filling. SUNCG \cite{song_semantic_2017} is a synthetic dataset of 3D scenes with realistic scene scale.
We render SUNCG to obtain a sequence of data by randomly sampling from SUN3D pose trajectories.
In addition to SUNCG and SUN3D, we generate a dataset --in the following called MVS-- with the COLMAP structure from motion pipeline \cite{schonberger_structure--motion_2016,schonberger_pixelwise_2016}.
MVS contains both indoor and outdoor scenes and was captured at full image and temporal resolution ($2704\times1520 @50 \text{Hz}$) with a wide-angle GoPro camera.
For training we downsample to ($320\times240$) and use every third frame.
We manually remove sequences where the reconstruction failed.

During training we use the (pseudo) ground truth camera poses from the datasets to construct the cost volume.

\subsubsection{Training Objective}
We use a simple L1 loss on the inverse depth maps $\mathcal{L}_\text{depth} = |\vect D - \vect D_\text{gt}|$ and the scale-invariant gradient loss proposed in \cite{ummenhofer_demon:_2017}:
\begin{equation}
	\mathcal{L}_\text{sc-inv-grad} = \sum_{h \in \{1,2,4\}} \sum_{i,j} \left\Vert \vect g_h[\vect D](i,j) - \vect g_h[\vect D_\text{gt}](i,j)\right\Vert_2,
\end{equation}
where
\begin{equation}
\vect g_h[\vect D](i,j) = \left( \tfrac{\vect D(i+h,j) - \vect D(i,j)}{\vert \vect D(i+h,j) \vert + \vert \vect D(i,j) \vert}, \tfrac{\vect D(i,j+h) - \vect D(i,j)}{\vert \vect D(i,j+h) \vert + \vert \vect D(i,j) \vert} \right)^\top.
\end{equation}
$\vect g_h[\vect D](i,j)$ and $\vect g_h[\vect D_\text{gt}](i,j)$ are gradient images of the predicted and the ground truth depth map that emphasize discontinuities.
$h$ is the step in the difference operator $\vect g_h$.

\section{Experiments}
\subsection{Tracking evaluation}
\tab{tab:tracking_error} shows the performance of our tracking network on the RGB-D benchmark \cite{sturm_benchmark_2012}.
The benchmark provides images and depth maps with accurate ground truth poses obtained from an external multi-camera tracking system.

We use the depth maps from the dataset during keyframe generation to measure the isolated tracking performance of our approach (left part of \tab{tab:tracking_error}).
We compare against the keyframe odometry component of the RGB-D SLAM method of Kerl~\etal~\cite{kerl_dense_2013}.
Their method uses the full color and depth information --for both keyframe and current frame-- to compute the pose, while our method only uses the depth information from the dataset for the keyframes.
During testing we generate a new keyframe if the rotational distance exceeds a threshold of 6 degrees or translational distance exceeds a 15cm threshold.
The number of generated keyframes is similar to the number of keyframes reported in \cite{kerl_dense_2013} for RGB-D SLAM.

\tab{tab:tracking_error} shows that our learning-based approach outperforms a state-of-the-art RGB-D method on most of the sequences, despite using less information.
In addition, the results also show that forcing the network to predict multiple pose hypotheses further reduces the translational drift on most sequences.
The results show also the generalization capabilities as we did not train or finetune on any sequences of the benchmark.

\begin{table}
\centering
\begin{tabular}{l||cccc|cc}
\toprule
\multicolumn{1}{c||}{} & \multicolumn{4}{c|}{Tracking} & \multicolumn{2}{c}{Tracking and mapping} \\
\midrule
\multirow{2}{*}{Sequence}       & RGB-D SLAM  & Ours     & Ours   & \multirow{2}{*}{Ours} & CNN-SLAM* & \multirow{2}{*}{Ours}             \\
      & Kerl~\etal~\cite{kerl_dense_2013} & (w/o flow) & (w/o hypotheses)  &               &  Tateno~\etal~\cite{tateno_cnn-slam_2017} & \\
\midrule
fr1/360      & 0.125     & 0.069 & 0.065                  & \textbf{0.054}            & 0.500              & \textbf{0.116}   \\
fr1/desk     & 0.037     & 0.042 & 0.031                  & \textbf{0.027}            & 0.095              & \textbf{0.078}   \\
fr1/desk2    & 0.020     & 0.025 & 0.020                  & \textbf{0.017}            & 0.115              & \textbf{0.055}   \\
fr1/plant    & 0.062     & 0.063 & 0.060                  & \textbf{0.057}            & \textbf{0.150}     & 0.165            \\
fr1/room     & 0.042     & 0.051 & 0.041                  & \textbf{0.039}            & 0.445              & \textbf{0.084}   \\
fr1/rpy      & 0.082     & 0.070 & \textbf{0.063}         & 0.065                     & 0.261              & \textbf{0.052}   \\
fr1/xzy      & 0.051     & 0.030 & 0.021                  & \textbf{0.019}            & 0.206              & \textbf{0.054}   \\
\midrule                  
average      & 0.060     & 0.050 & 0.043                  & \textbf{0.040 }            & 0.253               & \textbf{0.086 }   \\
\bottomrule
\end{tabular}
\caption{
Evaluation of our tracking (left part) and the combined mapping and tracking (right part) on the validation sets of RGB-D benchmark \cite{sturm_benchmark_2012}.
The values describe the translational RMSE in $[m/s]$.
\textbf{Tracking:} We compare the performance of our tracking network against the RGB-D SLAM method of Kerl~\etal~\cite{kerl_dense_2013}.
Numbers for Kerl~\etal~\cite{kerl_dense_2013} correspond to the frame-to-keyframe odometry evaluation and have been copied from their paper.
Kerl~\etal~\cite{kerl_dense_2013} uses the camera image \emph{and} the depth stream for computing the poses, while our approach uses the depth stream only for keyframes and is limited to photometric alignment. \textbf{Ours (w/o flow)} does not learn optical flow. \textbf{Ours (w/o hypotheses)} is a network which just predicts a single pose. \textbf{Ours} uses optical flow to learn motion features and predicts multiple pose hypotheses.
\textbf{Tracking and mapping:} We compare our tracking and mapping against CNN-SLAM by Tateno~\etal~\cite{tateno_cnn-slam_2017}.
\textbf{*} For a fair comparison CNN-SLAM is run without pose graph optimization.
To avoid a bias in the initialization \textbf{Ours} uses the depth prediction from CNN-SLAM for the first frame of each sequence and then switches to our combined tracking and mapping.
}
\label{tab:tracking_error}
\end{table}

\begin{table}
\centering
\scalebox{0.9}{
\begin{tabular}{l|c||ccc|ccc||cccc}
\toprule
\multicolumn{2}{c||}{}              &\multicolumn{3}{c|}{Fixed band} & \multicolumn{3}{c||}{Narrow band} & \multicolumn{4}{c}{Mapping comparison}\\
\midrule
\multicolumn{2}{c||}{}              & 2frames  & 6frames  & 10frames     &  1iter &  3iters  & 5iters  & SGM & DTAM & DeMoN & Ours\\
\midrule
 \multirow{3}{*}{MVS}
                      	& L1-inv & 0.117 & 0.085  & \bb{0.083}   &   0.076  & 0.065  & \bb{0.064}  &  -    & 0.086 & 0.059 & \bb{0.036}\\
                      	& L1-rel & 0.239 & 0.163  & \bb{0.159}   &   0.142  & 0.113  & \bb{0.111}  & -     & 0.557 & 0.240 & \bb{0.171}\\
                        & sc-inv & 0.193 & 0.160  & \bb{0.159}   &   0.156  & 0.132  & \bb{0.130}  & 0.251 & 0.305 & 0.246 & \bb{0.146}\\
\midrule
 \multirow{3}{*}{SUNCG}
                        & L1-inv & 0.075  & \bb{0.065}  & 0.067   &   0.049  & 0.039  & \bb{0.036} & -     & 0.142 & 0.169 & \bb{0.036}\\
                        & L1-rel & 0.439  & \bb{0.418}  & 0.423   &   0.304  & 0.213  & \bb{0.171} & -     & 0.380 & 0.533 & \bb{0.083}\\
                        & sc-inv & 0.213  & \bb{0.199}  & 0.200   &   0.174  & 0.152  & \bb{0.146} & 0.248 & 0.343 & 0.383 & \bb{0.128}\\
\midrule
 \multirow{3}{*}{SUN3D}
                        & L1-inv & 0.097  & 0.067 & \bb{0.065}    &   0.050  & \bb{0.035}  & 0.036 & -     & 0.210 & 0.197 & \bb{0.064}\\
                        & L1-rel & 0.288  & 0.198 &  \bb{0.193}   &   0.141  & \bb{0.082}  & 0.083 &  -    & 0.423 & 0.412 & \bb{0.111}\\
                        & sc-inv & 0.206  & 0.174 &  \bb{0.172}   &   0.155  & \bb{0.125}  & 0.128 & 0.146 & 0.374 & 0.340 & \bb{0.130}\\
\bottomrule
\end{tabular}
}
\vspace{1ex}
\caption{
Keyframe depth map errors on the test split of our training data sets.
\bb{Fixed band}: The influence of the number of frame used for computing the cost volume for the fixed band module.
Accumulating information from multiple frames improves the performance and saturates after adding six or more frames.
\bb{Narrow band}: The effect of different number of iterations of the narrow band module.
More iterations lead to more accurate depth maps.
Depth estimations converge after about three iterations and improve only slowly with more iterations.
On SUN3D results get slightly worse with more than three iterations.
The narrow band width $\sigma_{nb}$ is a constant number, which can be replaced by a gradually decreasing strategy or optimally by the uncertainty of the depth estimation. \bb{Mapping comparison}: Quantitative comparison to other learning- and cost-volume-based dense mapping methods.
We evaluate sequences of length 10 from our test sets and use the camera poses from the datasets to measure the isolated performance of our mapping.
DeMoN just uses two input images (first and last frame of each sequence) and does not use the pose as input.
Since DeMoN predicts the depth scaled with respect to its motion prediction, we compare only on the scale invariant metric sc-inv.
SUNCG and SUN3D feature a large number of indoor scenes with low texture, while MVS contains a mixture of indoor and outdoor scenes and provides more texture.
Our method outperforms the baselines on all datasets.
The margin is especially large on the very difficult indoor datasets (SUNCG, SUN3D).
}
\label{tbl:map_quant}
\end{table}

\subsection{Mapping evaluation}
For evaluating the mapping performance we use the following error metrics:
\begin{equation}
	\textstyle \text{sc-inv}(\vect D,\vect D_\text{gt}) = \sqrt{\tfrac{1}{n}\sum_{i,j} \vect E(i,j)^2 - \tfrac{1}{n^2} \left(\sum_{i,j} \vect E(i,j)\right)^2},
\end{equation}
where $ \vect E(i,j) = \log \vect D(i,j) - \log \vect D_\text{gt}(i,j)$ and $n$ is the number of pixels,
\begin{align}
	\text{L1-rel}(\vect D,\vect D_\text{gt}) &= \tfrac{1}{n} \textstyle{\sum_i} \frac{\vert \vect D(i,j) - \vect D_\text{gt}(i,j)\vert}{\vect D_\text{gt}(i,j)} \quad \text{and}\\
\text{L1-inv}(\vect D,\vect D_\text{gt}) &= \tfrac{1}{n} \textstyle{\sum_i} \left\vert \frac{1}{\vect D(i,j)} - \frac{1}{\vect D_\text{gt}(i,j)}\right\vert .
\end{align}
$\text{sc-inv}$ is a scale invariant metric introduced in \cite{eigen_depth_2014}.
The $\text{L1-rel}$ metric normalizes the depth error with respect to the ground truth depth value.
$\text{L1-inv}$ gives more importance to close depth values by computing the absolute difference of the reciprocal of the depth values.
This metric also reflects the increasing uncertainty in the depth computation with increasing distance to the camera.

\begin{figure}
\newcommand{\mywidth}{0.115\textwidth}
\centering
\begin{tabular}{c|| ccc |ccc ||c}
\toprule
                    & \multicolumn{3}{c|}{Fixed band} &\multicolumn{3}{c||}{Narrow band} & \\
\midrule
            Keyframe    & 2 frames& 6 frames& 10 frames &1 iter & 3 iters & 5 iters &GT \\
\midrule

    \includegraphics[width=\mywidth]{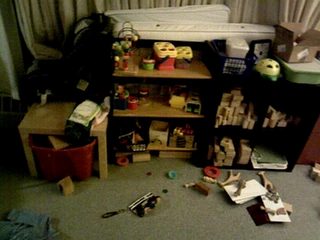}&
    \includegraphics[width=\mywidth]{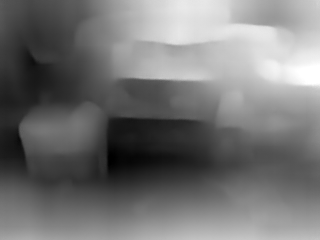} &
    \includegraphics[width=\mywidth]{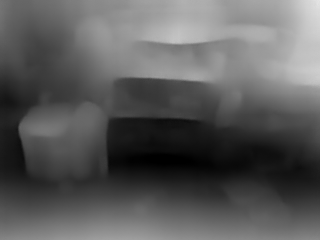}&
    \includegraphics[width=\mywidth]{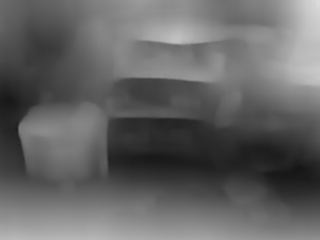}&
    \includegraphics[width=\mywidth]{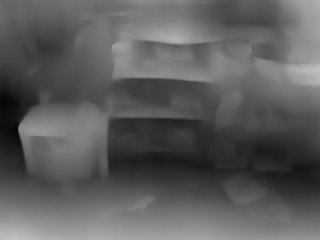}&
    \includegraphics[width=\mywidth]{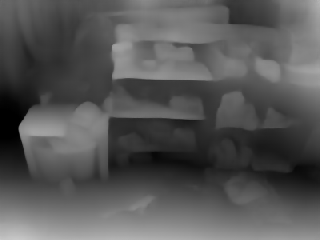}&
    \includegraphics[width=\mywidth]{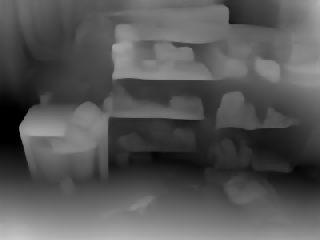}&
    \includegraphics[width=\mywidth]{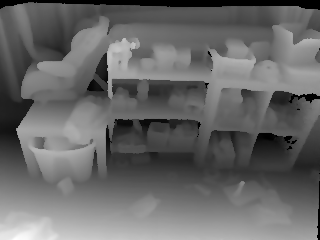}\\
    
    \includegraphics[width=\mywidth]{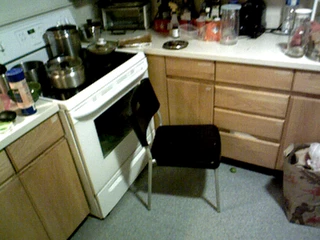}&
    \includegraphics[width=\mywidth]{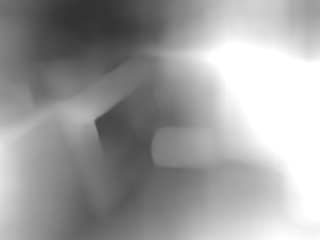}&
    \includegraphics[width=\mywidth]{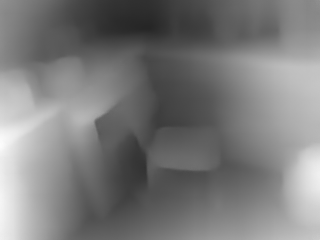}&
    \includegraphics[width=\mywidth]{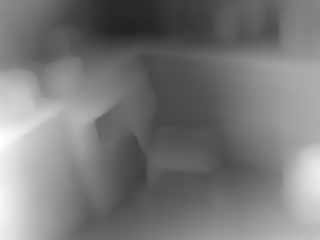}&
    \includegraphics[width=\mywidth]{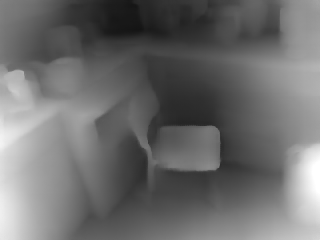}&
    \includegraphics[width=\mywidth]{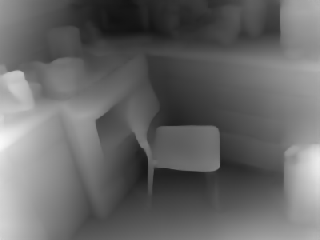}&
    \includegraphics[width=\mywidth]{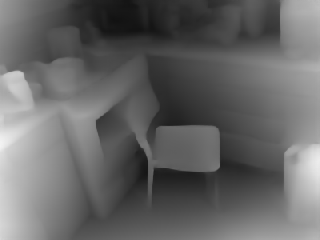}&
    \includegraphics[width=\mywidth]{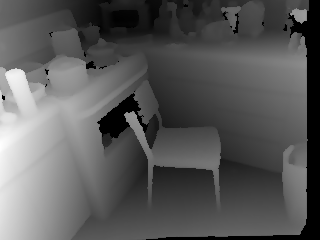} \\
    
    \includegraphics[width=\mywidth]{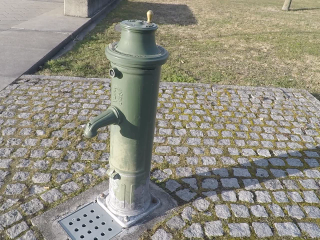}&
    \includegraphics[width=\mywidth]{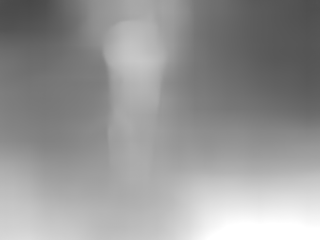}&
    \includegraphics[width=\mywidth]{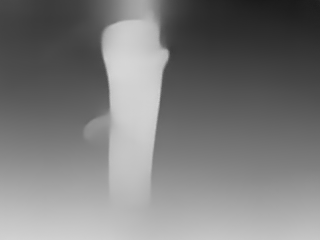}&
    \includegraphics[width=\mywidth]{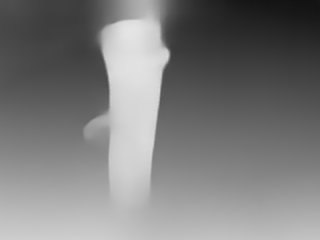}&
    \includegraphics[width=\mywidth]{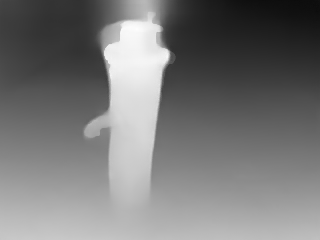}&
    \includegraphics[width=\mywidth]{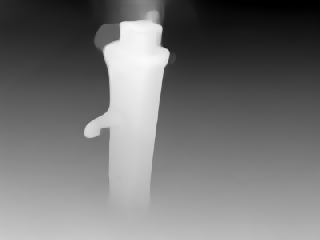}&
    \includegraphics[width=\mywidth]{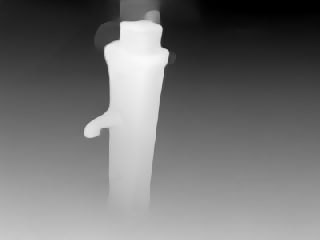}&
    \includegraphics[width=\mywidth]{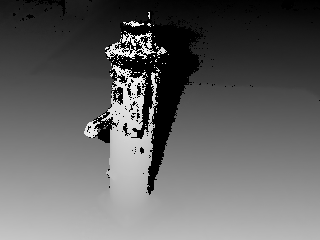}\\
\bottomrule
\end{tabular}
  \caption{%
Qualitative comparison of the depth prediction of the fixed band and narrow band module.
We evaluate the effect of different numbers of frames used in the fixed band module and iterations used in the narrow band module.
The fixed band gains in performance with more frames.
The largest improvement can be observed between using only 2 frames (including keyframe) and 6 frames.
The performance saturates with more frames.
To further improve the quality of the depth map we use the iterative narrow band module on the 10 frames result of the fixed band.
Using a narrow band around the previous depth estimation allows us to capture finer details and achieve higher accuracy.
}
\label{fig:fb_nb_performance}
\end{figure}

We evaluate our fixed band module and narrow band module quantitatively in \tab{tbl:map_quant}.
The results show that the fixed band module is able to exploit the accumulated information from multiple frames leading to better depth estimates.
While this behaviour is taken for granted for traditional methods, this is not necessarily the case for learning-based methods.
The same holds for iterative processes like the narrow band module.
Running the narrow band module iteratively improves the depth estimates.
We can show this quantitatively in \tab{tbl:map_quant} and qualitatively in \fig{fig:fb_nb_performance}.

We also compare our mapping against the state-of-the-art deep learning approach DeMoN \cite{ummenhofer_demon:_2017} and two strong classic dense mapping methods DTAM \cite{newcombe_dtam_2011} and SGM \cite{hirschmuller_accurate_2005}.
We use the publicly available reimplementation OpenDTAM\footnote{\url{https://github.com/magican/OpenDTAM.git} \tiny SHA: 1f92a54334c233f9c4ce7d8cbaf9a81dee5e69a6} and our own implementation of SGM with 16 directions.
For DTAM, SGM and DeepTAM we construct a cost volume with 32 labels at the resolution of $320\times240$.
We use SAD as photo-consistency measure and accumulate the information of video sequences of length 10.
We use the same pseudo camera pose ground truth from the datasets for a fair comparison.
For DeMoN --which is limited to two images-- we give the first and last frame from the sequence to provide enough motion parallax.

As shown in \tab{tbl:map_quant} our method achieves the best performance on all metrics and test sets.
All classic methods tend to suffer from weakly-textured scenes which occur quite often in the indoor datasets and synthetic datasets.
However, we are less affected by this problem by means of leveraging matching cost information together with scene priors via a neural network.
This is again supported by the qualitative comparison in \fig{fig:mapping_comparison}.
In addition, the mapping performance of all the classic cost-volume-based methods is prone to noisy camera pose while our method is more robust, which is demonstrated in \fig{fig:mapping_noise_comparison}.
More qualitative examples can be found in the supplemental video.

In the right part of \tab{tab:tracking_error} we compare our combined tracking and mapping against CNN-SLAM~\cite{tateno_cnn-slam_2017} without pose graph optimization. CNN-SLAM uses a semi-dense photoconsistency optimization approach for computing camera poses and uncertainty-based depth update. We did not train on RGB-D benchmark datasets \cite{sturm_benchmark_2012}. Our learned dense tracking and mapping generalizes well and proves to be more robust and accurate on the majority of sequences.
While it performs clearly worse on fr1/plant it seldom fails and overall yields more reliable trajectories.

To further verify our generalization ability, we test our model on KITTI \cite{geiger_are_2012} without finetuning. \fig{fig:kitti} shows a qualitative comparison.

\begin{figure}
  \centering
\begin{tikzpicture}[every node/.style={inner sep=0, outer sep=0}]
	\matrix[column sep=1mm, row sep=0mm] (mymatrix)
	{
		\node{}; &
		\node{Keyframe}; &
		\node{SGM}; &
		\node{DTAM}; &
		\node{DeMoN}; &
		\node{Ours}; &
		\node{GT}; \\
		
		
		\node[rotate=90]{SUN3D}; &
		\node{\includegraphics[width=0.15\textwidth]{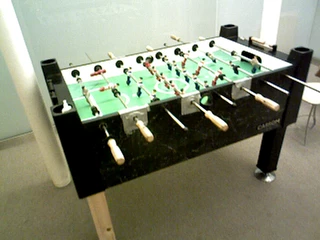}}; &
		\node{\includegraphics[width=0.15\textwidth]{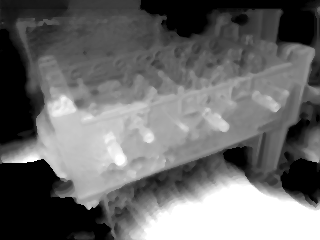}}; &
		\node{\includegraphics[width=0.15\textwidth]{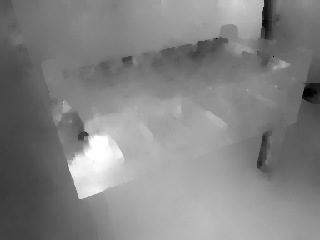}}; &
		\node{\includegraphics[width=0.15\textwidth]{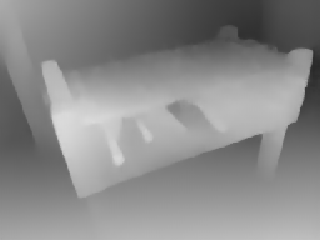}}; &
		\node{\includegraphics[width=0.15\textwidth]{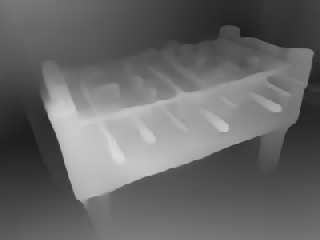}}; &
		\node{\includegraphics[width=0.15\textwidth]{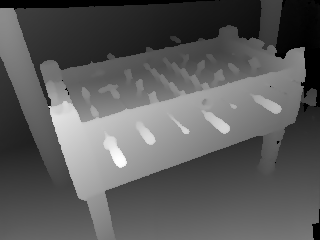}}; \\

		\node[rotate=90]{SUNCG}; &
		\node{\includegraphics[width=0.15\textwidth]{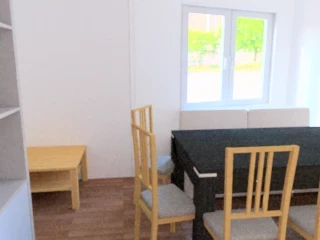}}; &
		\node{\includegraphics[width=0.15\textwidth]{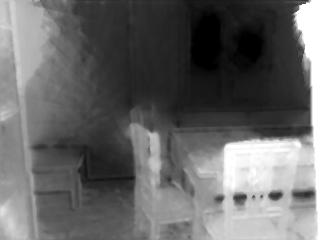}}; &
		\node{\includegraphics[width=0.15\textwidth]{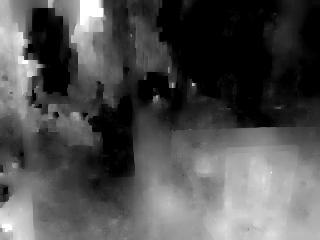}}; &
		\node{\includegraphics[width=0.15\textwidth]{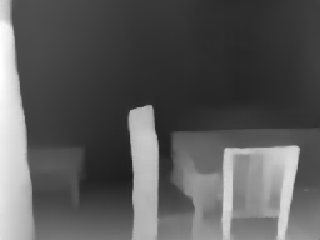}}; &
		\node{\includegraphics[width=0.15\textwidth]{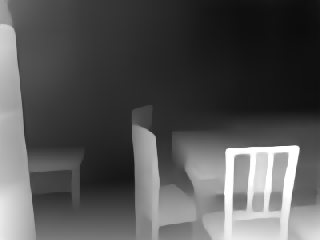}}; &
		\node{\includegraphics[width=0.15\textwidth]{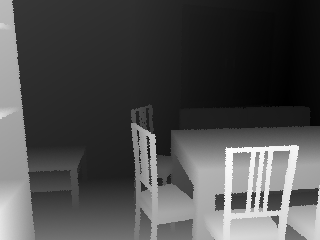}}; \\

		\node[rotate=90]{MVS}; &
		\node{\includegraphics[width=0.15\textwidth]{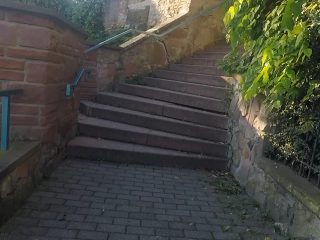}}; &
		\node{\includegraphics[width=0.15\textwidth]{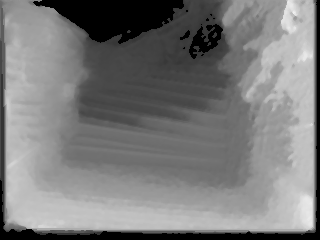}}; &
		\node{\includegraphics[width=0.15\textwidth]{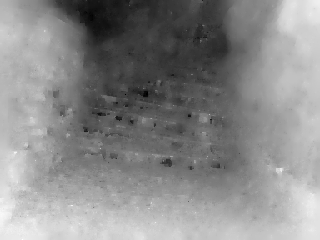}}; &
		\node{\includegraphics[width=0.15\textwidth]{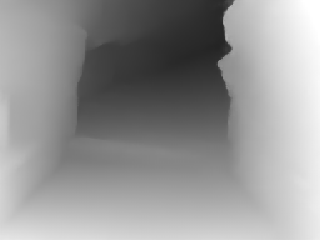}}; &
		\node{\includegraphics[width=0.15\textwidth]{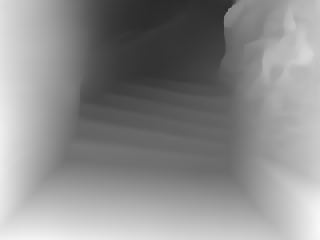}}; &
		\node{\includegraphics[width=0.15\textwidth]{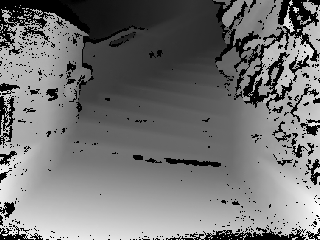}}; \\
	};
\end{tikzpicture}
  \caption{%
Qualitative depth prediction comparison for sequences with 10 frames.
DTAM has problems with short sequences and textureless scenes.
SGM shares the same problems but works reasonably well if enough texture is present.
DeMoN work well even in homogeneous image regions but misses many details.
Our method can produce high quality depth maps using a small number of frames and captures more details compared to the other methods.
}
  \label{fig:mapping_comparison}
\end{figure}
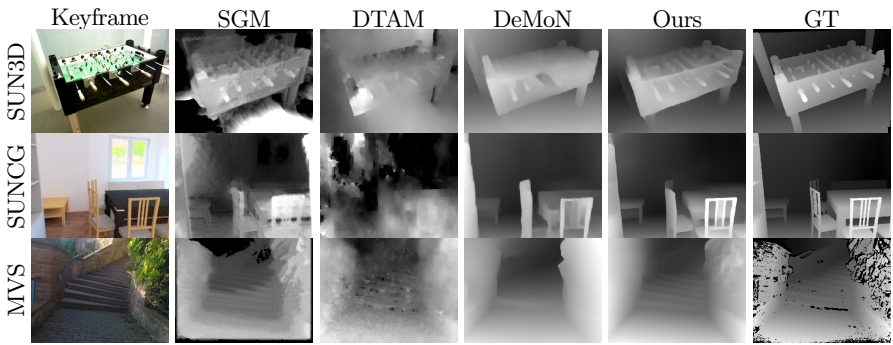

\begin{figure}
  \centering
    \incgraphicsbrbox{width=0.30\textwidth}{kitti/seq98_img_0.png}{Image}
    \incgraphicsbrbox{width=0.30\textwidth}{kitti/seq98_dgm_pr.png}{SGM}  
    \incgraphicsbrbox{width=0.30\textwidth}{kitti/seq98_dtam_pr.png}{DTAM} 
    
    \incgraphicsbrbox{width=0.30\textwidth}{kitti/seq98_demon_pr.png}{DeMoN} 
    \incgraphicsbrbox{width=0.30\textwidth}{kitti/seq98_nb_4.png}{Ours} 
    \incgraphicsbrbox{width=0.30\textwidth}{kitti/seq98_gt.png}{GT} 
  \caption{%
  Generalization experiment on KITTI \cite{geiger_are_2012}. SGM, DTAM and Ours use a sequence of 5 frames from the left color camera, while for DeMoN we only use the first and last frame of each sequence. We show pseudo GT as a reference, which was obtained by computing the disparity of the corresponding rectified and synchronized stereo pairs. KITTI is an urban scene dataset captured with a wide-angle camera, which differs from our training data significantly. 
Further, due to the dominant forward motion pattern of the dataset the epipole is within the visible image borders, which makes depth estimation especially difficult.
Without finetuning our method generalizes well to this dataset. More examples can be found in the supplementary.
  }
  \label{fig:kitti}
\end{figure}

\begin{figure}
  \centering
\begin{tikzpicture}[every node/.style={inner sep=0, outer sep=0}]
	\matrix[column sep=1mm, row sep=1mm] (mymatrix)
	{

		\node[rotate=90]{SGM}; &
		\node{\includegraphics[width=0.13\textwidth]{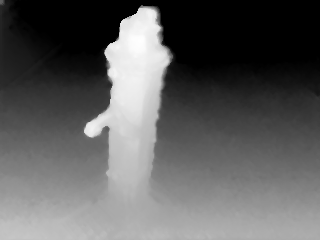}}; &
		\node{\includegraphics[width=0.13\textwidth]{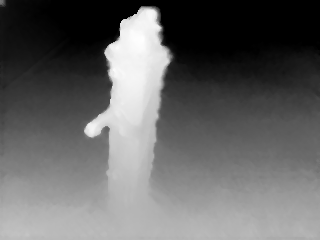}}; &
		\node{\includegraphics[width=0.13\textwidth]{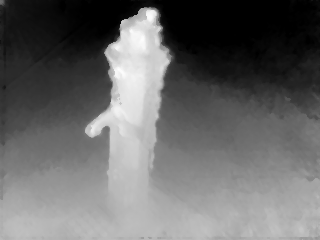}}; &
		\node{\includegraphics[width=0.13\textwidth]{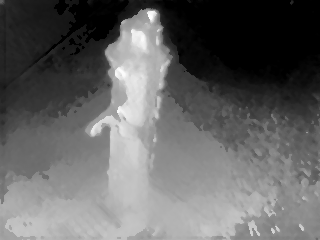}}; &
		\node{\includegraphics[width=0.13\textwidth]{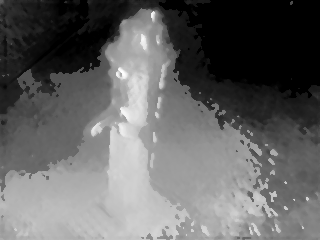}}; &
		\node{\includegraphics[width=0.13\textwidth]{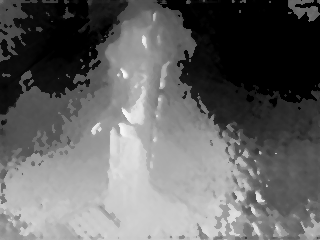}}; &
		\node{\includegraphics[width=0.13\textwidth]{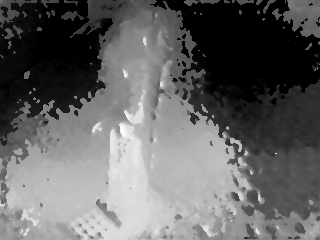}}; \\
		\node[rotate=90]{DTAM}; &
		\node{\includegraphics[width=0.13\textwidth]{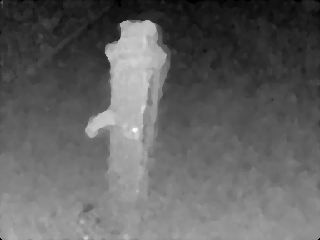}}; &
		\node{\includegraphics[width=0.13\textwidth]{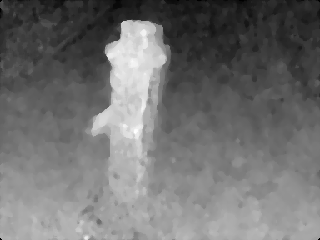}}; &
		\node{\includegraphics[width=0.13\textwidth]{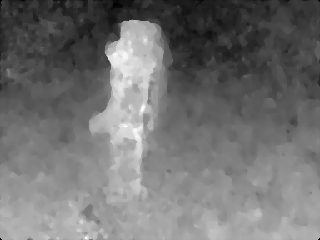}}; &
		\node{\includegraphics[width=0.13\textwidth]{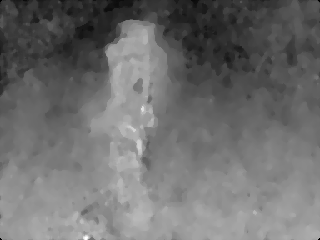}}; &
		\node{\includegraphics[width=0.13\textwidth]{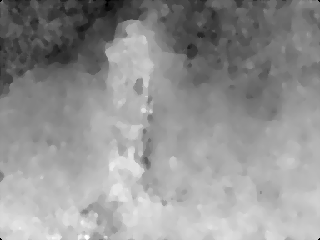}}; &
		\node{\includegraphics[width=0.13\textwidth]{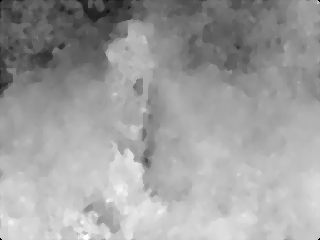}}; &
		\node{\includegraphics[width=0.13\textwidth]{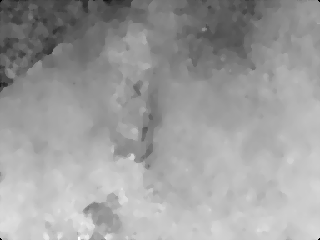}}; \\
		
                \node[rotate=90]{Ours}; &
		\node{\includegraphics[width=0.13\textwidth]{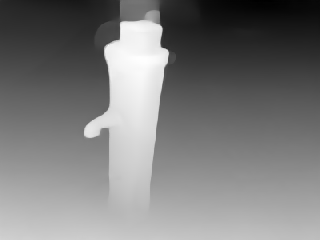}}; &
		\node{\includegraphics[width=0.13\textwidth]{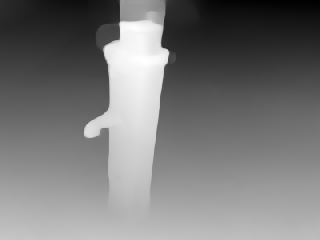}}; &
		\node{\includegraphics[width=0.13\textwidth]{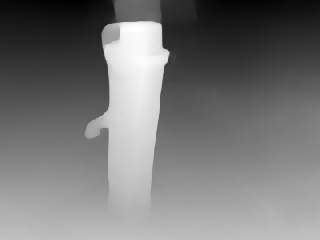}}; &
		\node{\includegraphics[width=0.13\textwidth]{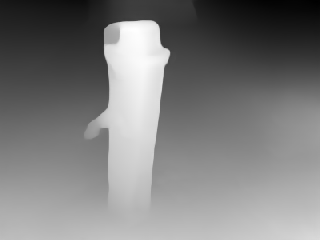}}; &
		\node{\includegraphics[width=0.13\textwidth]{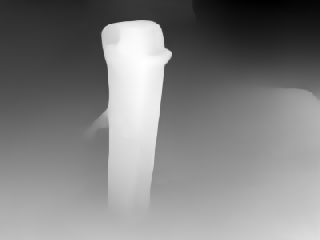}}; &
		\node{\includegraphics[width=0.13\textwidth]{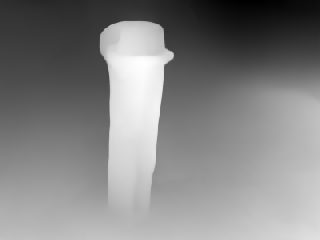}}; &
		\node{\includegraphics[width=0.13\textwidth]{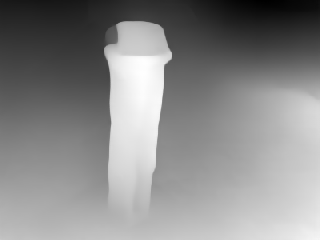}}; \\
	};
	\draw[very thick,->] (-5,-2.2) --node[below=1mm]{pose noise} (5,-2.2);
\end{tikzpicture}
  \caption{%
Qualitative depth prediction comparison of DeepTAM, SGM, DTAM against increasing pose noise.
We carefully select a well textured video sequence with 10 frames and enough motion parallax.
For SGM and DTAM we use a cost volume with 64 labels, while we use 32 labels for DeepTAM.
We found that using 64 instead 32 labels improves the results for both baseline methods.
We apply the same normal-distributed noise vectors for all methods to the camera poses and increase the standard deviation from $0$ (leftmost) to $0.6 |\vect \xi|$ (rightmost).
SGM and DTAM are highly sensitive to noise and their performance degrades quickly.
Our predicted depth preserves the important scene structures even under large amounts of noise.
This behaviour is advantageous during tracking and improves the robustness of the overall system.
}
  \label{fig:mapping_noise_comparison}
\end{figure}
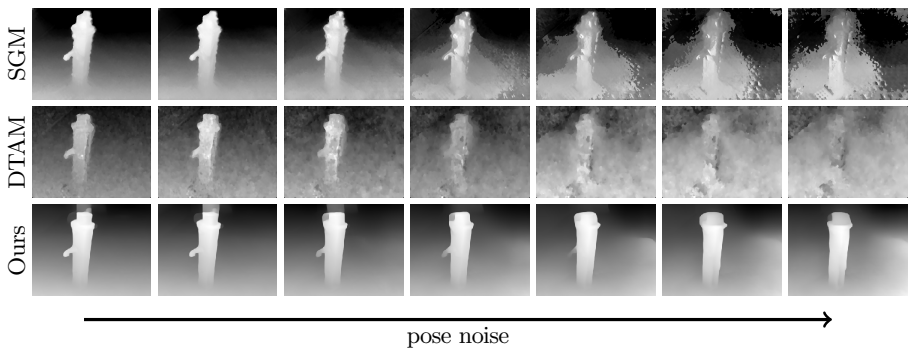

\section{Conclusions}
We propose a novel deep learning architecture for real-time dense mapping and tracking.
For tracking we show that generating synthetic viewpoints allows us to track incrementally with respect to a keyframe.
For mapping, our methods can effectively exploit the cost volume information and image-based priors leading to accurate and robust dense depth estimations.
We demonstrate that our methods outperform strong classic and deep learning algorithms. In future work, we plan to extend the presented components to build a full SLAM system.\\[1ex]
\noindent\textbf{Acknowledgements}\quad This project was in large parts funded by the EU Horizon 2020 project Trimbot2020. 
We also thank the bwHPC initiative for computing resources, Facebook for their P100 server donation and gift funding.

\clearpage

\bibliographystyle{splncs04}
\bibliography{2996}

\clearpage
\includepdf[pages=-]{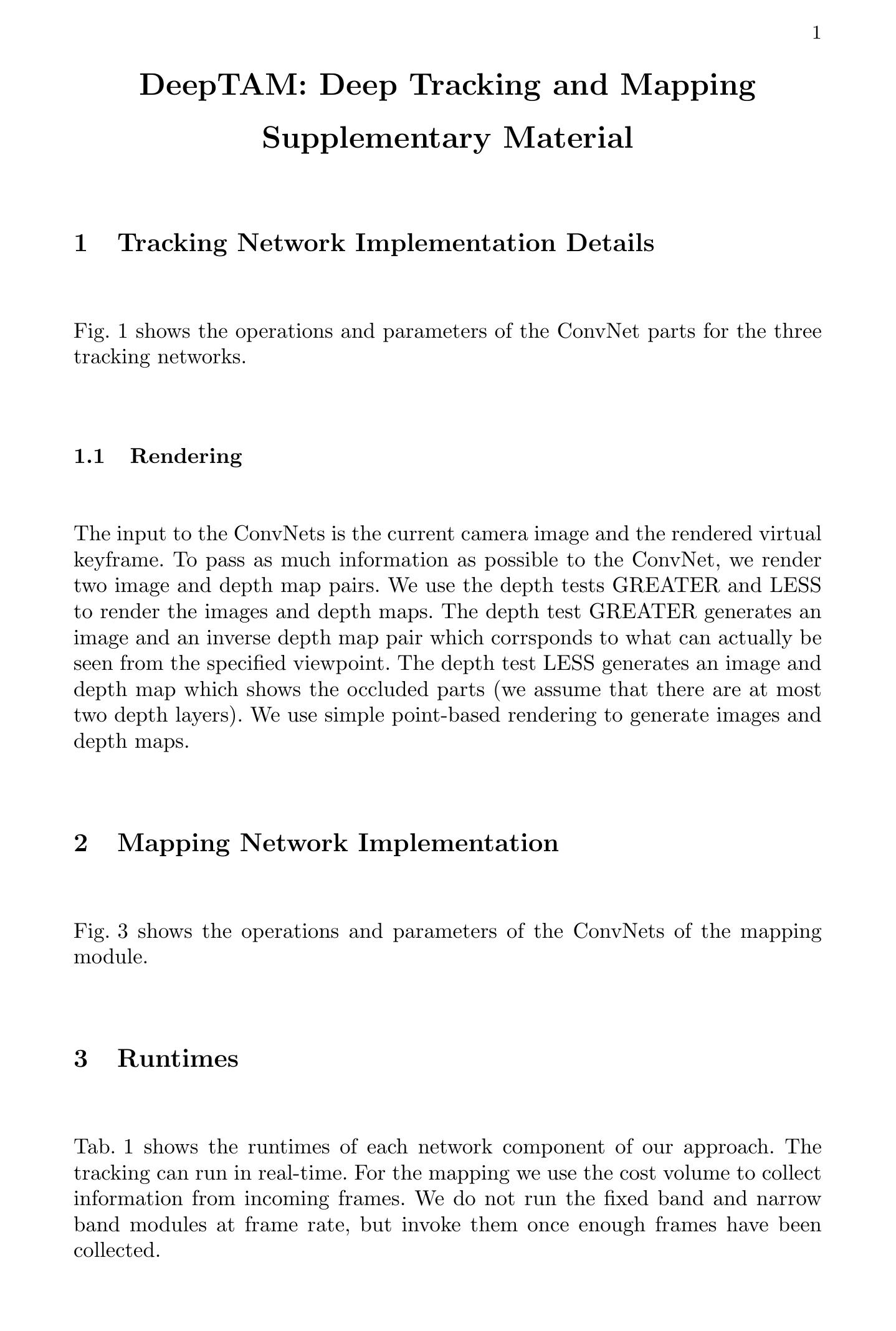}
\end{document}